\documentclass[runningheads]{llncs}

\usepackage{eccv}

\usepackage{eccvabbrv}
\usepackage{soul}
\usepackage{graphicx}
\usepackage{booktabs}
\usepackage{caption}
\usepackage{bbding} 
\usepackage{afterpage}
\usepackage{bbding}
\usepackage[accsupp]{axessibility}  

\usepackage{hyperref}

\usepackage{orcidlink}

\begin{document}

\title{Pondering the Way: Spatial-perceiving World Action Model for Embodied Navigation
} 




\titlerunning{Spatial-perceiving World Action Model for Navigation}

\author{Hong Chen\inst{1}\orcidlink{0000-0003-1110-713X}\textsuperscript{(\Envelope)} \and Daqi Liu\inst{2}\thanks{Project leaders.}\orcidlink{0000-0003-1929-657X}\textsuperscript{(\Envelope)} \and Zehan Zhang\inst{2}\protect\footnotemark[1]\textsuperscript{(\Envelope)} \and Haiguang Wang\inst{2} \and Tianhao Lu\inst{1} \and Longfei Yan\inst{3} \and Haiyang Sun\inst{2} \and Fangzhen Li\inst{2} \and Hongwei Xie\inst{2} \and Bing Wang\inst{2} \and Guang Chen\inst{2} \and Hangjun Ye\inst{2}\thanks{Corresponding authors.}\textsuperscript{(\Envelope)} \and Yihua Tan\inst{1}\protect\footnotemark[2]\orcidlink{0000-0003-0963-5339}\textsuperscript{(\Envelope)}  } 



\authorrunning{Chen et al.}

\institute{
Huazhong University of Science and Technology, Wuhan, China \\
\email{\{hongc, yhtan\}@hust.edu.cn} 
\and
Xiaomi EV, Beijing, China \\
\email{liudaqikk@gmail.com, zehanzhang@126.com, yehangjun@gmail.com} 
\and
Zhejiang University, Hangzhou, China \\
}

\maketitle
\noindent\begin{center}
\includegraphics[width=\textwidth]{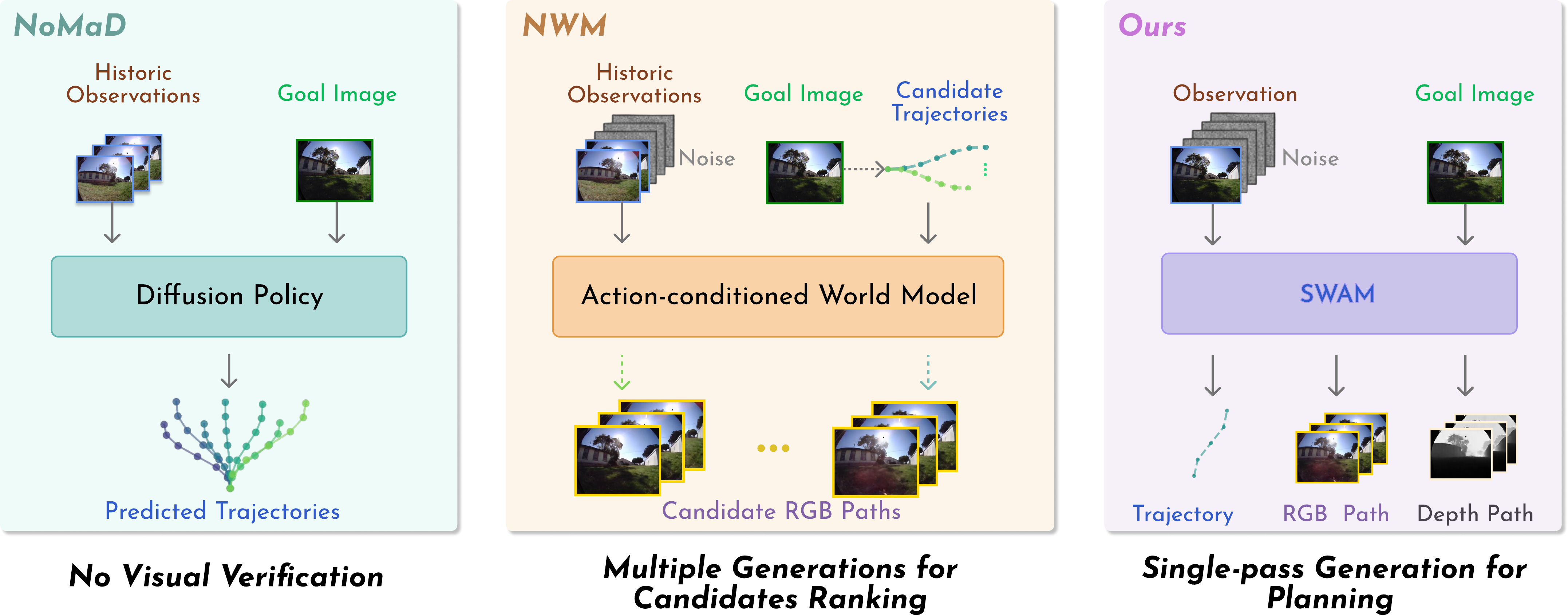}\\[0.2cm]
 \captionof{figure}{Comparison of different paradigms for visual navigation. \textbf{Left:} Nomad predicts trajectories from historical RGB frames and a goal image. \textbf{Middle:} NWM generates future RGB videos conditioned on trajectories. \textbf{Right (Ours):} Our model jointly predicts trajectories and RGBD paths from the current RGB observation and the goal image in one forward pass, enabling efficient planning.}
\label{fig:te}
\end{center}

\begin{abstract}
Existing world model-based planners for visual navigation typically follow a verification-centric paradigm, decoupling goal intent from trajectory synthesis. This approach suffers from candidate dependence, heavy computational overhead, and inconsistencies between sampled actions and predicted visuals. To address these issues, we propose SWAM (\textbf{S}patial-perceiving \textbf{W}orld \textbf{A}ction \textbf{M}odel), a task-centric joint observation-action generation framework. Given start and goal RGB observations, SWAM performs single-pass inference to simultaneously generate intermediate RGB-D sequences and corresponding action trajectories, promoting goal-consistent trajectory generation and improved spatial feasibility. While SWAM leverages depth pseudo-labels during training to internalize spatial priors, it requires only monocular RGB input at inference time. We further introduce a visual-guided action refinement module and a trajectory-scale regularization loss to enforce fine-grained alignment between motion and visual cues while stabilizing predictions across varying distances. Extensive experiments show that SWAM significantly outperforms state-of-the-art two-stage planners in success rate, trajectory accuracy, and inference efficiency, while demonstrating robust zero-shot generalization to unseen environments.
\end{abstract}

\section{Introduction}
\label{sec:intro}

Visual goal-conditioned navigation ~\cite{bonin2008visual,zhang2022survey} is a core challenge for intelligent systems operating in the physical world. The task demands that an agent, given a visual goal and its current observation, must infer where to move next, which paths are traversable, and when the goal has been successfully reached. This requires tight integration of visual perception, spatial reasoning, and decision-making \cite{DeepVisual, hafner2020dreamer}. 

Direct online policy methods learn observation-to-action mappings~\cite{shah2021ving, shah2023gnm, shah2023vint, sridhar2024nomad, team2024octo, wu2025uniphys}. While inference-efficient, they lack explicit visual horizons and struggle with error recovery. Offline world model-based planners~\cite{bar2025navigation,yang2025mindjourney} adopt a two-stage pipeline: sample candidate action sequences from an external policy~\cite{sridhar2024nomad,chi2023diffusionpolicy}, rollout visual trajectories via an action-conditioned video predictor, and select the trajectory best matching the goal. While this enables simulation-evidence-based evaluation, its modular design constrains the world model to verification rather than trajectory synthesis, leading to three limitations: (i) candidate dependence, decision quality is bounded by candidate coverage; (ii)  computational inefficiency, exhaustive sampling and rollout incur prohibitive inference costs, especially in complex or longer planning settings; (iii) geometric inconsistency, absence of explicit spatial constraints can produce abrupt viewpoint changes or infeasible trajectories, reducing the reliability of simulation evidence. 

We argue that these limitations stem from decoupling the path proposal and environment prediction. Verification-centric pipelines treat the world model~\cite{wang2025mila,yang2025resim,shah2023navigation} as a passive evaluator, decoupling goal intent from trajectory generation and relegating spatial reasoning to implicit visual patterns, which neglects the core requirement of navigation. 

To address this, we propose a Spatial-perceiving World Action Model (SWAM), a joint observation–action generation framework. Given a start RGB observation and a goal RGB observation, our model performs single-pass inference to jointly generate an intermediate RGBD observation sequence and the corresponding action trajectory. Unlike unidirectional world models~\cite{bar2025navigation, zhou2025learning}, our approach co-generates "how to act" and "what to perceive" within a unified process. This approach jointly constrains actions and observations throughout generation, ensuring goal alignment, temporal consistency, and spatial feasibility. Unlike verification-centric world models, SWAM directly synthesizes control trajectories through joint latent denoising. SWAM offers three key advantages: (1) goal alignment is maintained throughout generation, producing structured, target-directed paths; (2) inference cost is reduced by eliminating candidate expansion and repeated evaluation; (3) explicit injection of spatial cues enhances the constraints on actions.

Technically, we leverage DepthAnything v3~\cite{lin2025depth} to provide per-frame depth pseudo-labels during training, leveraging spatial priors while requiring only monocular RGB at test time. Estimated depth supplies sufficient navigable cues, avoiding reliance on scarce ground-truth depth. We further introduce a Visual-Guided Action Refinement (VGAR) module and Trajectory-Scale Regularization (TSR) loss to enhance the alignment between predicted actions and generated visual cues, and to stabilize trajectory predictions over various distances.

Experiments show that SWAM outperforms strong policy-based and two-stage baselines in trajectory error, goal success rate, visual fidelity, and practical inference efficiency, while demonstrating more stable long-distance and cross-dataset behavior. 
In conclusion, our contributions are threefold: 
\begin{itemize}
\item We introduce SWAM, a joint observation–action generation framework for visual navigation, replacing candidate-based pipelines; 
\item We propose a visual-guided action refinement and trajectory-scale regularization loss to  better leverage visual cues for better action prediction and reduce trajectory drift error;
\item We validate SWAM across multiple datasets, showing improved accuracy, success rate, visual quality, inference efficiency, and robust cross-scene generalization.
\end{itemize}

\section{Related Work}
\label{sec:related}

\subsection{Visual Goal-Conditioned  Navigation}
Visual goal-conditioned navigation aims to learn policies that guide an agent to reach a target location specified by a goal observation. Early approaches formulate this problem as learning a mapping from the current observation and a goal image to control actions, typically using reinforcement learning or imitation learning \cite{zhu2017target, savinov2018semi, gupta2017cognitive}. These methods learn deterministic policies that directly predict actions from visual observations and have demonstrated strong performance in simulated indoor environments. Subsequent work improves policy learning through representation learning and memory mechanisms, enabling agents to reason over partial observations and long-horizon dependencies \cite{mirowski2017learning, parisotto2018neural}. However, deterministic policy learning often struggles with the multi-modality of navigation behaviors, where multiple valid trajectories can reach the same goal. To address this limitation, recent works explore generative formulations of navigation policies. In particular, diffusion-based policies have been proposed to model distributions over action trajectories, enabling more diverse and robust navigation strategies \cite{sridhar2024nomad, chi2023diffusionpolicy}. While these approaches improve the expressiveness of goal-conditioned policies, they typically generate action sequences alone, without explicitly modeling the intermediate visual states along the trajectory. In contrast, our approach formulates navigation as joint video–action generation conditioned on the goal observation, allowing the model to explicitly capture the visual transitions connecting the start and goal states.

\subsection{Action-Conditioned World Models for Planning and Navigation}
World models~\cite{allen1983planning,hao2023reasoning,li2025comprehensive} aim to learn predictive models of environment dynamics that allow agents to plan by imagining future observations under candidate actions. Early works, such as visual model predictive control, predict future images conditioned on candidate action sequences and optimize actions by minimizing perceptual distance to a goal observation \cite{hirose2019deep, ebert2018visual}. Recent advances in generative modeling~\cite{ho2020denoising,lipmanflow,songdenoising} have significantly improved the capacity of world models. A number of works adopt video prediction or video diffusion models to model environment dynamics for planning \cite{Ha2018WorldM}. More recently, large-scale generative models have been used to construct action-conditioned video diffusion world models~\cite{liu2024sora,wan2025wan,zheng2024open} capable of generating realistic imagined rollouts \cite{bar2025navigation, yang2025mindjourney, zhou2025learning, zhang2025epona,gao2026dreamdojo }. These predicted trajectories can then be used for planning via trajectory ranking or search. Despite their strong generative ability, these pipelines generally follow an action-first, verify-later paradigm, where candidate actions are first sampled from policies or search algorithms and then evaluated using the world model. As a result, planning performance depends heavily on the coverage of candidate trajectories and the available rollout budget, which can become computationally expensive for long-horizon navigation tasks. In contrast, our method directly generates coherent action–observation trajectories conditioned on the goal in a single generative process, avoiding candidate trajectory sampling and improving planning efficiency.

\subsection{Joint Video Action Modeling for Embodied Intelligence}

Recent advances in embodied AI emphasize the joint modeling of observations and actions to capture the causal dynamics of agent-environment interactions. By unifying these modalities within generative world models, researchers have significantly improved agent planning, controllability, and generalization.

Navigation-oriented world models have transitioned from basic future observation predictors to spatially grounded reasoning systems. Models such as DINO-WM \cite{zhou2024dinowm} and WoMaP \cite{yin2025womap} leverage pretrained visual features and structured modeling to enhance open-vocabulary localization and zero-shot planning. Further scaling perception-action integration, World-in-World \cite{zhang2025worldinworld} and UniDrive-WM \cite{xiong2026unidrivewm} explore closed-loop learning and unified perception-planning in driving scenarios. However, these methods primarily treat the world model as an environment simulator, often decoupling environment dynamics from the explicit generation of executable action sequences.

A parallel trend focuses on joint video-action generation, where models like VideoVLA \cite{shenvideovla}, CoVAR \cite{yang2025covar}, and others \cite{kim2024openvla, black2024pi_0, cen2025worldvla } demonstrate that unified architectures can excel in robotic manipulation. Recent works such as World Action Models \cite{ye2026world} shows that action-aware world models can serve as zero-shot policies. While PAD \cite{pad2024} shares our goal of joint modeling, these existing approaches predominantly target short-horizon manipulation or local workspace interactions. They frequently lack the long-horizon spatiotemporal priors and explicit geometric grounding essential for complex navigation tasks. More recently, Aether~\cite{zhu2025aether} combines geometric reconstruction and generative modeling to improve spatial understanding and prediction. Although navigation-related scenarios are included, navigation-specific planning metrics and goal-conditioned trajectory generation are not explicitly evaluated.

In contrast to the aforementioned research, SWAM addresses various-distance planning through a unified perception-action generative process with spatial perceiving.
\section{Method}
\label{sec:method}
\begin{figure}[t]
    \centering
    \includegraphics[width=1.0\textwidth]{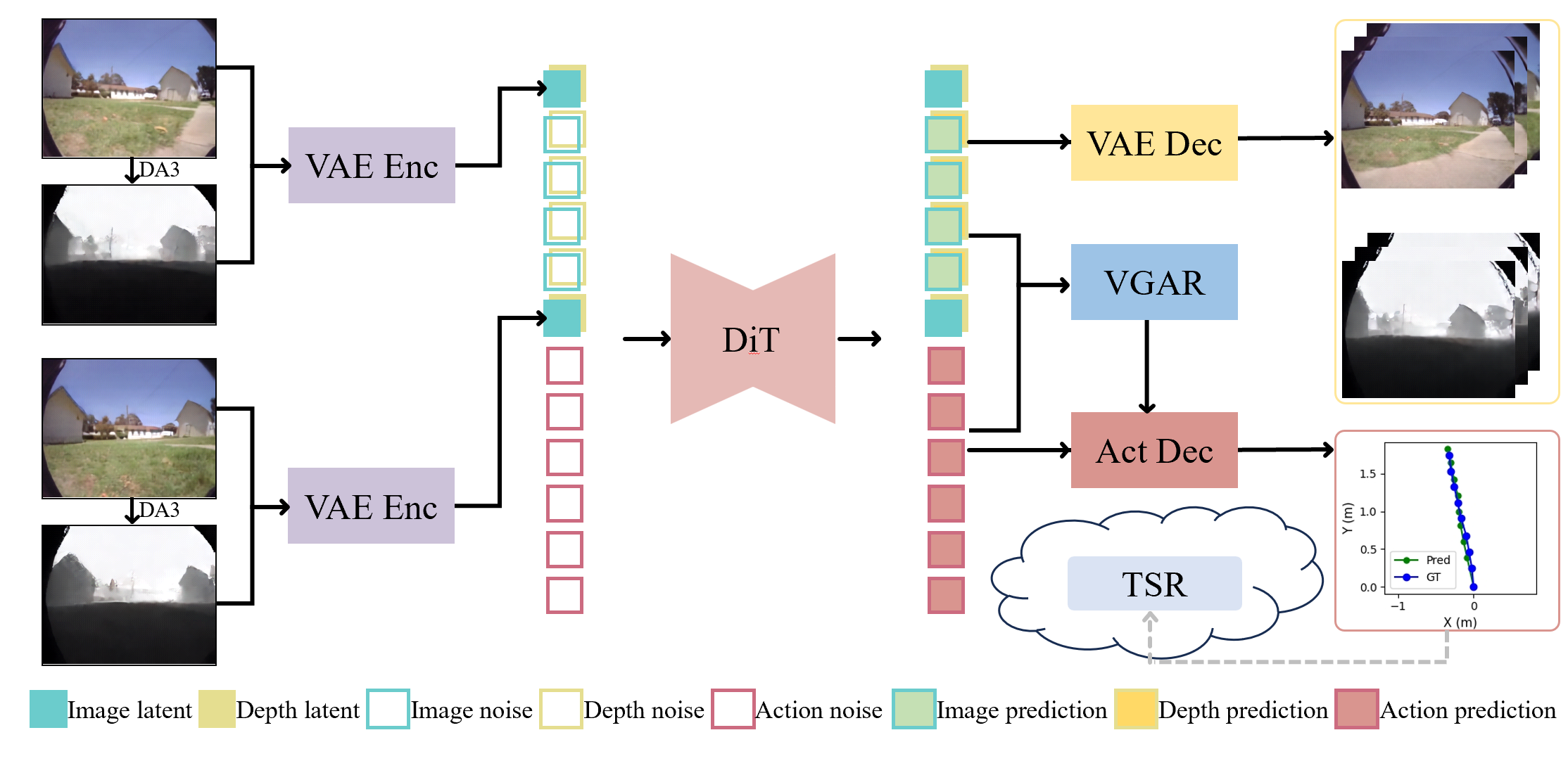}
    \caption{Overview of SWAM. We extend the pretrained Diffusion Transformer (DiT) of CogVideoX \cite{yang2025cogvideox} by conditioning it on start and goal frame latents, and fine-tune it to jointly produce intermediate RGB-D frame latents and associated action tokens. DepthAnything V3~\cite{lin2025depth} is used to predict pseudo-depth maps for the start and goal frames. The Visual-Guided Action Refinement (VGAR) module further refines the predicted actions via cross-attention with RGBD latents. During training, losses are imposed both on the diffusion denoising process and our proposed Trajectory-Scale Regularization (TSR) loss to ensure local plausibility and global trajectory consistency.}
    \label{fig:architecture}
    \vspace{-3mm}
\end{figure}

In this section, we present the proposed SWAM framework for goal-conditioned visual navigation, including the unified observation-action generation architecture, the Visual-Guided Action Refinement (VGAR) module, the Trajectory-Scale Regularization (TSR) loss, and the training strategy built upon a pretrained video diffusion backbone. More detailed information is shown in the Appendix.

\subsection{Problem Formulation}
\label{sec:problem_formulation}
We consider goal-conditioned visual navigation, where an agent is given a start observation and a goal observation, and is required to generate a feasible trajectory connecting them. Let $o_0=(I_0, D_0)$ and $ o_G=(I_G, D_G)$ denote the start and goal observations, where (I) represents an RGB image and (D) denotes the corresponding depth representation.

Our objective is to jointly generate a horizon-(N) sequence of future observations ($\{o_n\}_{n=1}^{N}$) and actions ($\{a_n\}_{n=1}^{N}$, where $a_n=(\Delta x_n,\Delta y_n)$) denotes the planar motion of the robot in its local coordinate frame. Specifically, we learn the conditional distribution 
$$
p_\theta\left(\{o_n\}_{n=1}^{N},\{a_n\}_{n=1}^{N}\mid o_0,o_G\right).
$$
During training, depth representations are obtained using an off-the-shelf monocular depth estimator and are used as auxiliary geometric supervision to provide spatial grounding for the generative process.

\subsection{Overall Architecture}
\label{sec:architecture}

Fig.~\ref{fig:architecture} illustrates the overall framework. We adopt a latent diffusion model that jointly generates (i) an RGBD observation sequence and (ii) the corresponding action sequence. This joint modeling encourages temporal-spatial consistency: actions are optimized to be compatible with the predicted visual evolution, and the generated visual evolution is conditioned on the action plan.

\noindent\textbf{Latent Representation and Tokenization.} Given each observation denoted as $o_n=(I_n,\hat{D}_n)$, we employ the frozen 3D variational autoencoder (VAE) of CogvideoX with spatial-temporal convolutions to encode the RGB image and depth map into latent feature maps, separately:
RGB latent  $z_n^{i}\in\mathbb{R}^{h\times w\times d}$ and depth latent $z_n^{d}\in\mathbb{R}^{h\times w\times d}$, where $h\times w$ is the latent spatial resolution and $d$ is the feature dimension. We represent an action sequence $\{a_n\}_{n=1}^{N}$ as a token matrix $x^{a}\in\mathbb{R}^{N\times D}$ by projecting each 2D action into the same embedding dimension using a learnable MLP. This enables unified modeling over visual and action tokens.

 \noindent\textbf{Multimodal Sequence Construction.}
We construct one unified token sequence that contains conditions and denoised targets. Specifically, we define the clean target variables (to be diffused) as: intermediate RGB tokens $\{z_n^{i}\}_{n=1}^{N}$, intermediate depth tokens $\{z_n^{d}\}_{n=1}^{N}$, and action tokens $x^{a}$. We concatenate them into a single sequence $X_t$
and treat the start/goal latents as conditioning information rather than part of $X_t$:
$
c = (z_0^i,z_0^d,z_G^i,z_G^d)
$, where $t$ represents the denoising step.




\noindent\textbf{Denoising Process.} We adopt a DDPM-style forward noising process on the unified sequence. In the denoising step $t$, the diffusion transformers (DiTs) \cite{peebles2023scalable}  $\mathcal{F}_\theta$ iteratively denoise the sequence over $T$ steps. At denoising step $t$, the model predicts the noise residual $\epsilon_\theta$ given the current noisy sequence $X_t$ and goal conditioning:
\begin{equation}
    X_{t-1} = \frac{1}{\sqrt{\alpha_t}} \left( X_t - \frac{1 - \alpha_t}{\sqrt{1 - \bar{\alpha}_t}} \epsilon_\theta(X_t, z^i_0, z^d_0, z^i_G, z^d_G, t) \right) + \sigma_t z,
\end{equation}
where $\alpha_t$ and $\bar{\alpha}_t$ are diffusion schedule parameters, $\sigma_t$ is the sampling variance, and $z \sim \mathcal{N}(0, I)$ (omitted when $t = 0$). After denoising completes, RGB and depth latents are decoded to image space using the VAE decoder, while action tokens are projected to planar actions through a lightweight network $f$ with the proposed VGAR module as described in Sec~\ref{sec:depth_guide}.


\subsection{Visual-Guided Action Refinement} \label{sec:depth_guide} Since the final action readout is less sensitive to local geometric cues, we introduce a lightweight Visual-Guided Action Refinement (VGAR) module that injects generated RGBD evidence into action tokens right before decoding, without modifying the joint diffusion backbone. Let $X_v$ and $X_a$ denote the final-layer visual and action tokens from the diffusion Transformer. VGAR refines $X_a$ via a gated residual cross-attention: \begin{gather} C = \text{CrossAttn}\!\big(\text{LN}(X_a),\; \text{LN}(X_v),\; \text{LN}(X_v)\big), \\ \Delta X_a = G \odot W(C), \quad G = \sigma\!\big(\text{MLP}([\text{LN}(X_a);\, C])\big), \\ X_a’ = X_a + \Delta X_a, \end{gather} where $\text{LN}$ is LayerNorm, $[\cdot;\cdot]$ denotes feature concatenation, $W$ is a learnable projection, and $\sigma$ is the sigmoid function. The gate $G$ controls how much visual evidence flows into the action representation. An action head then decodes the refined tokens $X_a’$ into action sequences.

\subsection{Joint-Training Objective} \label{sec:goal_losses} \subsubsection{Diffusion Loss.} We train with the standard DDPM denoising objective~\cite{ho2020denoising, songscore}: \begin{equation} \mathcal{L}_{\text{DDPM}} = \mathbb{E}_{t, X_0, \epsilon} \left[ \left\| \epsilon - \epsilon_\theta\!\left(\sqrt{\bar{\alpha}_t}\, X_0 + \sqrt{1 - \bar{\alpha}_t}\, \epsilon,\; \mathbf{c},\; t\right) \right\|_2^2 \right], \end{equation} where $\mathbf{c} = (z^i_0, z^d_0, z^i_G, z^d_G)$ denotes the conditioning set (image, depth, and goal tokens), $X_0$ is the clean latent sequence (RGB, depth, and action tokens), and $\bar{\alpha}_t = \prod_{s=0}^{t} \alpha_s$ is the cumulative noise schedule. \subsubsection{Trajectory-Scale Regularization.} We observe that diffusion loss alone on actions leads to compounding error when actions are integrated into a trajectory, with drift growing more severe over long horizons. To counteract this, we introduce a Trajectory-Scale Regularization (TSR) loss that directly supervises the cumulative displacement: \begin{equation} \mathcal{L}_\text{TSR} = \frac{\| \hat{p}_N - p_G \|_2}{N}, \quad \hat{p}_N = \sum_{n=1}^{N} \hat{a}_n, \end{equation} where $\hat{p}_N$ is the predicted endpoint obtained by integrating denoised actions and $p_G$ is the ground-truth goal position in the local coordinate frame. The $1/N$ normalization ensures stability across trajectories of varying length. \subsubsection{Training Objective.} The final objective combines diffusion denoising with the trajectory regularizer: \begin{equation} \mathcal{L} = \mathcal{L}_{\text{DDPM}} + \lambda_{\text{TSR}}\, \mathcal{L}_{\text{TSR}}. \end{equation} Among sequences with similar local denoising error, TSR favors those that preserve the correct global displacement, turning long-distance drift into a directly learnable signal. This improves stability and generalization under varying trajectory lengths and limited-data regimes.


\subsection{Model Initialization and Architectural Adaptation}
\label{sec:training}

To preserve the rich spatiotemporal priors of the pretrained video generation model while avoiding distribution shifts caused by newly introduced parameters, we initialize all newly added modules with zero weights.  This strategy ensures that, at the beginning of training, the model behaves identically to the original pretrained generator and produces high-quality RGB predictions based on its learned priors. During training, gradients gradually update the new parameters, allowing the model to progressively learn the coupling between RGB observations, depth signals, and actions in a data-driven manner.  Such progressive adaptation stabilizes training and enables seamless integration of additional modalities without disrupting the pretrained feature space.

\section{Experiments}
In this section, we present a comprehensive evaluation of our Spatial-perceiving World Action Model (SWAM). We first describe the experimental setup in Section~\ref{sec:exp_setup}, including datasets, baseline methods, and evaluation metrics. Section~\ref{sec:main_results} presents the main quantitative results, followed by zero-shot generalization performance in Section~\ref{sec:generalization}, demonstrating the superiority of our proposed framework. We then provide qualitative analysis in Section~\ref{sec:qualitative} to illustrate the visual and trajectory consistency of our approach (additional visualizations are provided in the Appendix). Finally, Section~\ref{sec:ablation} conducts ablation studies to validate the effectiveness of key components in our framework.

\label{sec:experiment}


\subsection{Experimental Setup}
\label{sec:exp_setup}

\textbf{Datasets.}
We conduct a comprehensive evaluation of our method on four benchmark navigation datasets. RECON \cite{recon} comprises large-scale outdoor trajectories with extended horizons, thereby testing the model’s capacity for long-term planning in complex real-world environments. SCAND \cite{scand} encompasses both indoor and outdoor scenes with dynamic obstacles and social agents, presenting challenges in reasoning about real-time interactions and heterogeneous scene structures. TartanDrive \cite{tartandrive} evaluates performance in off-road driving scenarios characterized by pronounced forward-motion biases and irregular terrain, which necessitates adaptation to constrained and non-uniform action distributions. Finally, HuRoN \cite{huron} provides an indoor setting with dynamic human interactions, and we utilize it for evaluating the zero-shot generalization to previously unseen social dynamics. For all datasets, action signals are derived from consecutive robot poses and normalized to a unified scale across different embodiments, ensuring comparability of policy outputs across diverse navigation contexts. We follow the same data preprocessing, trajectory segmentation, and evaluation protocols established by NWM \cite{bar2025navigation} during evaluation to ensure methodological consistency and fair comparison.

\noindent{\textbf{Baselines.}}
We evaluate our approach against representative direct policy methods, including GNM \cite{shah2023gnm} and the diffusion-based navigation policy NoMaD \cite{sridhar2024nomad}. Additionally, we consider world-model-based planners by integrating NWM \cite{bar2025navigation} with NoMaD \cite{sridhar2024nomad} to generate candidate trajectories, which are subsequently ranked based on perceptual similarity. We evaluate NWM using varying numbers of sampled candidate trajectories ($N \in \{2, 4, 8, 16\}$) for ranking to demonstrate how candidate diversity influences final planning performance. This ensures the $\times N$ notation reflects the inference-time sampling budget rather than model scaling. 

Since SWAM is initialized from the publicly available CogVideoX  \cite{yang2025cogvideox} backbone, we additionally construct a strong CogVideoX-based baseline to isolate the contribution of the proposed navigation-specific designs for fair comparisons. Specifically, we extend CogVideoX to jointly generate future observations and action sequences under the same training protocol, conditioning strategy, and pretrained initialization used by SWAM. The baseline predicts future RGB observations and corresponding actions conditioned on the start and goal frames, but does not incorporate depth modeling, trajectory-scale regularization, or visual-guided action refinement. This baseline enables an ablation-style evaluation, isolating the contribution of our proposed components while maintaining a consistent backbone architecture or pretraining.

\noindent{\textbf{Evaluation Metrics.}} We evaluate navigation performance along three complementary dimensions: trajectory accuracy, goal-reaching success, and video generation quality. Unless otherwise specified or indicated for the meters metric, trajectory accuracy is quantified using Absolute Trajectory Error (ATE) and Relative Pose Error (RPE), with all metrics reported in unit grids for consistency with NWM \cite{bar2025navigation}. Goal-reaching performance is measured via success@($\tau$) at multiple distance thresholds ($\tau \in {1.0, 0.5, 0.25}$ grids), capturing both coarse and fine-grained navigation success rate. To assess the quality of the generated video paths, we compute PSNR, SSIM, and LPIPS between predicted and ground-truth RGB frames. All metrics are evaluated on the same standardized subsets as NWM, and reported results correspond to the average over three independent runs. Inference time is measured per episode without parallelization, providing a realistic assessment of runtime efficiency.


\subsection{Main Results}
\label{sec:main_results}
We report the main experimental results that validate the effectiveness of our SWAM approach. Further related results can be found in the appendix.

\begin{table*}[t]
\caption{Trajectory error (ATE/RPE) across datasets. Lower values are better. \textbf{Bold} indicates best performance; \underline{underline} indicates second best.}
\label{tab2}
\centering
\resizebox{0.9\textwidth}{!}{
\begin{tabular}{l|cc|cc|cc|c}
\toprule
Method & \multicolumn{2}{c|}{RECON} & \multicolumn{2}{c|}{SCAND} & \multicolumn{2}{c|}{TartanDrive} &Time(s)\\
& ATE & RPE & ATE & RPE & ATE & RPE   \\
\midrule
GNM & 1.85 & 0.54 & \underline{2.18} & 0.61 & 6.68 & 1.63  &0.12   \\
NoMaD & 1.90 & 0.53 & 2.36 & 0.52 & 6.66 & 1.36 &0.21\\
NWM+NoMaD ($\times$2) & 1.87 & 0.52 & 2.41 & 0.51 & 6.45 & 1.32 &31.37\\
NWM+NoMaD ($\times$4) & 1.82 & 0.52 & 2.42 & 0.49 & 6.40 & 1.32 	&63.83\\
NWM+NoMaD ($\times$8) & 1.74 & 0.50 & 2.37 & 0.48 & 6.31 & 1.31 &118.32\\
NWM+NoMaD ($\times$16) & \underline{1.53} & \underline{0.49} & \underline{2.18} & \underline{0.46} & 6.23 & 1.30 &245.98\\
CogVideoX (Joint) & 2.09 & 0.73 & 2.25 & 0.67 & \underline{4.90} & \underline{1.07} &14.12 \\
\midrule
\textbf{Ours} & \textbf{0.93} & \textbf{0.43} & \textbf{1.15} & \textbf{0.34} & \textbf{1.55} & \textbf{0.68} 	&16.91\\
\bottomrule
\end{tabular}
}
\end{table*}

\noindent\textbf{Trajectory Accuracy.} Tab.~\ref{tab2} shows ATE/RPE across datasets. SWAM achieves substantially lower trajectory errors than all baselines—outperforming NWM + NoMaD ($\times$16) by 39.2\% (RECON), 47.2\% (SCAND), and 75.1\% (TartanDrive) in ATE. Notably, SWAM requires only single-pass inference (16.91s/sample), while NWM+NoMaD ($\times$16) requires exhaustive rollout (245.98s/sample), demonstrating superior efficiency–accuracy tradeoff. Notably, CogVideoX achieves the second-best ATE on TartanDrive, significantly outperforming NWM-based methods. These results indicate that joint RGB-action generation provides useful cues for trajectory prediction. However, CogVideoX's performance remains far inferior to our full model (1.55), and its RPE is notably worse (1.07 vs. 0.68), indicating that visual coherence alone is insufficient for precise navigation. Our approach builds upon this insight by explicitly coupling action prediction with RGBD generation and spatial guidance, yielding both geometrically feasible trajectories and visually consistent rollouts.

\begin{figure}[t]
    \centering
    \includegraphics[width=0.45\textwidth]{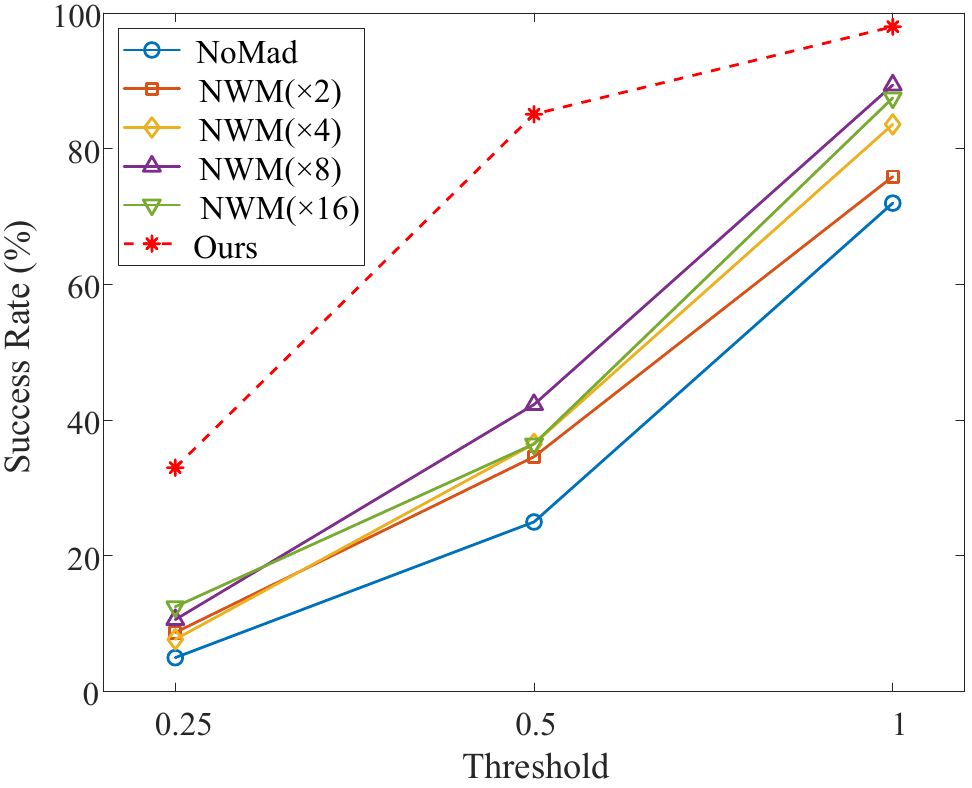}
    \caption{Success@($\tau$) curves across thresholds on RECON. SWAM achieves substantially higher success rates, especially at strict thresholds ($\tau=0.25$ and $\tau=0.5$).}
    \label{Exp_3}
\end{figure}

\noindent\textbf{Goal-reaching Performance.} Fig.~\ref{Exp_3} reports success@($\tau$) curves across thresholds. SWAM shows particularly strong gains at strict thresholds ($\tau=0.25$), achieving a 2.1$\times$ higher success than NWM+NoMaD ($\times$16) on RECON. This indicates that SWAM enhances final-positional precision, which closely aligns with navigation planning.

\noindent\textbf{Video Generation Quality.} Tab.~\ref{tab1} summarizes video generation metrics.  SWAM achieves state-of-the-art results across all datasets. We attribute the improved visual quality to two factors. First, spatial grounding via predicted depth provides strong geometric constraints that regularize the generation process, reducing visual artifacts and improving structural consistency across frames. This is particularly evident on TartanDrive, where our method achieves the highest PSNR (18.11) and SSIM (0.532), indicating that depth-aware generation produces more realistic terrain and obstacle appearances. Second, more accurate action prediction directly translates to better observation generation: since our actions are geometrically feasible and closely aligned with ground-truth trajectories, the corresponding RGBD frames naturally exhibit higher fidelity to the actual observations. This coupling ensures that improved action accuracy provides more realistic visual conditioning, which in turn refines future action predictions.
\begin{figure}[tbp]
    \centering
    \includegraphics[width=1.0\textwidth]{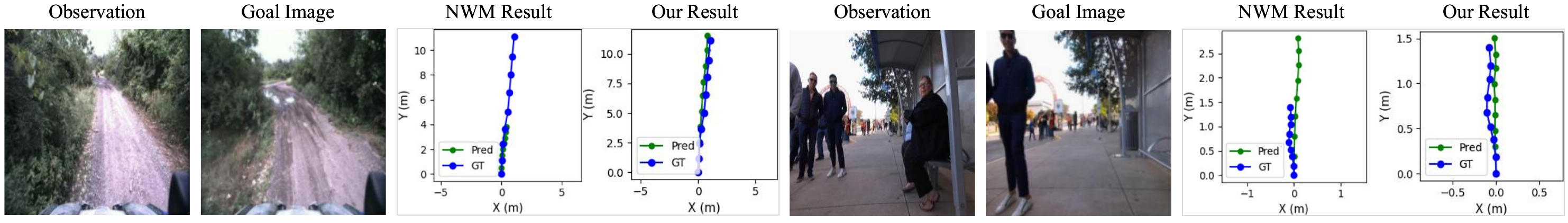}
    \caption{Robustness to path length. SWAM maintains accurate trajectory scaling across distances, while NWM exhibits obvious trajectory scale errors.}
    \label{Exp2}
\end{figure}

\begin{figure}[ht]
    \centering
    \includegraphics[width=0.92\textwidth]{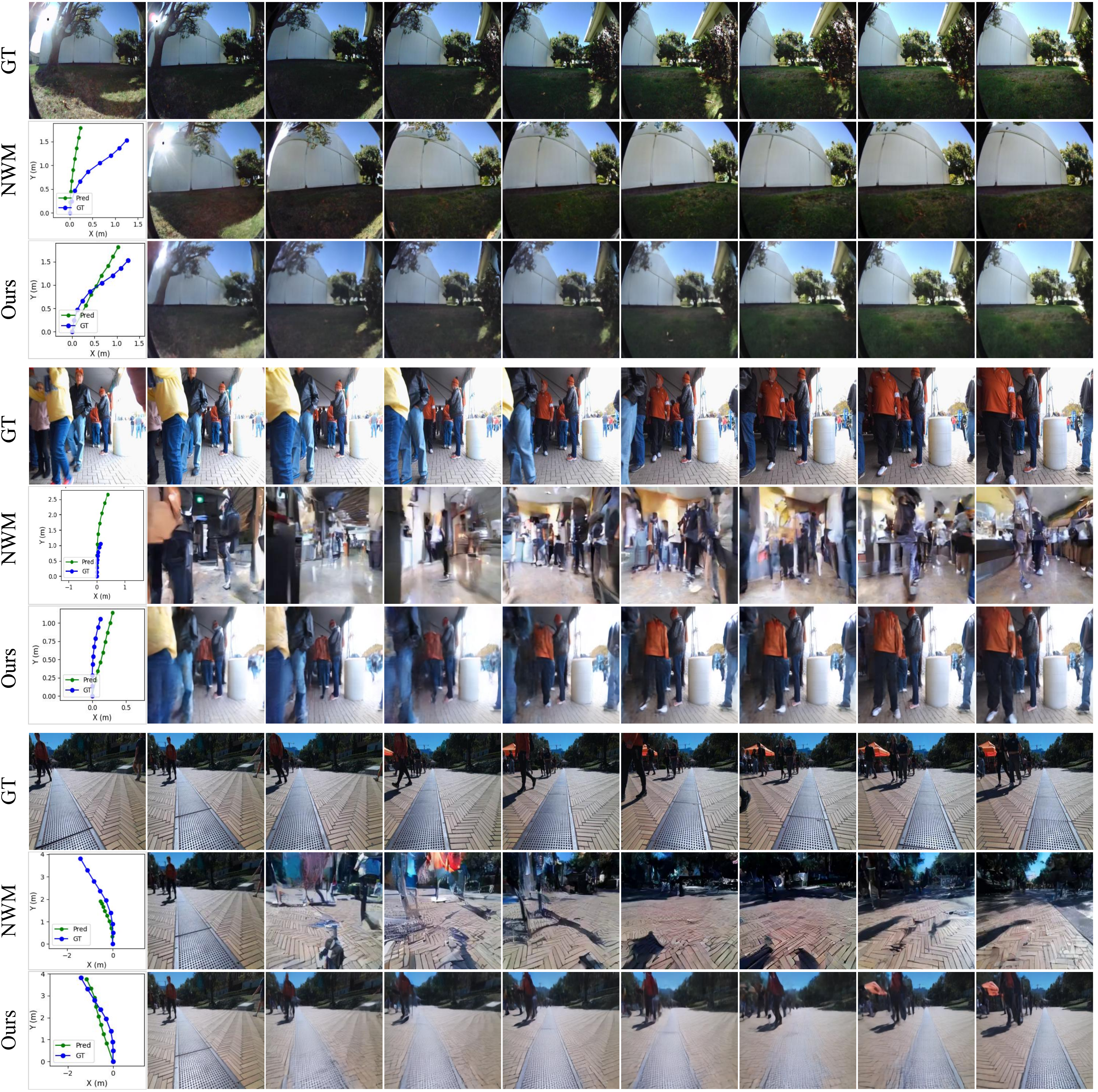}
    \caption{Qualitative comparison between SWAM and NWM. SWAM generates trajectories closer to ground truth with consistent video generation, while NWM struggles in such challenging scenarios.}
    \label{Exp1}
\end{figure}

 \begin{table*}[t]
\caption{Video generation (PSNR/SSIM/LPIPS). Higher PSNR/SSIM and lower LPIPS are better. \textbf{Bold} indicates best performance; \underline{underline} indicates second best.}
\label{tab1}
\centering
\resizebox{0.95\textwidth}{!}{
\begin{tabular}{l|ccc|ccc|ccc}
\toprule
Method & \multicolumn{3}{c|}{RECON} & \multicolumn{3}{c|}{SCAND} & \multicolumn{3}{c}{TartanDrive} \\
& PSNR$\uparrow$ & SSIM$\uparrow$ & LPIPS$\downarrow$ & PSNR$\uparrow$ & SSIM$\uparrow$ & LPIPS$\downarrow$ & PSNR$\uparrow$ & SSIM$\uparrow$ & LPIPS$\downarrow$ \\
\midrule
NWM+NoMaD ($\times$2) & 11.84 & 0.360 & 0.470 & 11.66 & 0.360 & 0.476 & 13.75 & 0.276 & 0.402 \\
NWM+NoMaD ($\times$4) & 14.14 & 0.435 & 0.364 & 11.85 & 0.367 & 0.468 & 13.84 & 0.277 & 0.427 \\
NWM+NoMaD ($\times$8) & 14.47 & 0.439 & 0.351 & 11.79 & 0.363 & 0.470 & 13.64 & 0.278 & 0.390 \\
NWM+NoMaD ($\times$16) & 14.45 & 0.438 & \underline{0.350} & 11.95 & 0.368 & 0.471 & 13.22 & 0.284 & 0.378 \\
CogVideoX (Joint) & \underline{16.65} & \underline{0.633} & 0.366 & \underline{15.04} & \underline{0.601} & \underline{0.351} & \underline{17.20} & \underline{0.524} & \textbf{0.330} \\
\midrule
\textbf{Ours} & \textbf{17.31} & \textbf{0.653} & \textbf{0.328} & \textbf{16.11} & \textbf{0.619} & \textbf{0.325} & \textbf{18.11} & \textbf{0.532} & \underline{0.335} \\
\bottomrule
\end{tabular}
}
\end{table*}

 \subsection{Qualitative Analysis} \label{sec:qualitative} We provide qualitative analysis across different scenes and motion conditions. For NWM, we visualize the trajectory with the best trajectory ranking score to ensure a fair comparison. Fig.~\ref{Exp2} evaluates robustness under nearly straight-line motion with varying trajectory lengths. SWAM maintains accurate trajectory scaling across distances (e.g., $\sim$10\,m and 1.5\,m in two examples), whereas NWM exhibits scale drift that grows with the prediction horizon, indicating that SWAM effectively mitigates scale accumulation errors over varying distances. Fig.~\ref{Exp1} compares trajectory and video generation across diverse cases. SWAM correctly captures turning trajectories where NWM shows a strong straight-motion bias. In complex open environments, SWAM preserves both trajectory accuracy and spatiotemporal consistency in generated observations, while NWM gradually loses contextual information, leading to mode collapse, hallucinated frames, and large trajectory errors. Overall, SWAM produces trajectories that closely follow the ground truth with minimal drift while maintaining temporally coherent video generation, whereas NWM struggles to jointly maintain geometric accuracy and visual consistency in challenging scenarios. These results demonstrate that SWAM better preserves the coupling between action dynamics and visual observations, leading to more reliable predictions.

\begin{table*}[t]
\caption{Zero-shot generalization to HuRoN (ATE/RPE). Lower values are better. \textbf{Bold} indicates best performance.}
\label{tab4}
\centering
\resizebox{0.65\textwidth}{!}{
\begin{tabular}{l|cc}
\toprule
Method & ATE & RPE \\
\midrule
NWM+NoMaD ($\times$16) (trained on HuRoN) & 3.73 & 0.95 \\
\midrule
\textbf{Ours} (zero-shot, no HuRoN training) & \textbf{2.94} & \textbf{0.85} \\
\bottomrule
\end{tabular}}
\end{table*}

\subsection{Zero-Shot Generalization}
\label{sec:generalization}

We further evaluate the generalization ability of SWAM under unseen environments. For this evaluation, we conduct zero-shot transfer experiments on the HuRoN dataset without any training or fine-tuning. Tab.~\ref{tab4} shows the results. Despite never being trained on HuRoN, our model achieves an ATE of 2.94 and an RPE of 0.85, i.e., 21.2\% lower ATE than NWM+NoMaD ($\times$16), which was explicitly trained on HuRoN. This result demonstrates the strong cross-domain generalization capability of SWAM across environments with different layouts and visual characteristics.

\subsection{Ablation Studies}

\label{sec:ablation}

To analyze the contribution of each component in our framework, we conduct ablation studies on three datasets: RECON, SCAND, and TartanDrive, using CogVideoX as the baseline. This model supports joint video-action generation, allowing us to fairly evaluate each enhancement while keeping the generative backbone consistent. Tab.~\ref{tab5} reports trajectory accuracy (ATE/RPE) and video generation quality (PSNR), providing a comprehensive view of how each module affects navigation precision and visual fidelity.

\noindent{\textbf{Effect of Additional Depth Modality.}}
Adding depth modality provides explicit spatial perception cues for video generation and trajectory prediction. With depth incorporated, trajectory accuracy improves significantly over the baseline. On RECON, ATE decreases from 2.09 to 1.70 and RPE from 0.70 to 0.47. Similar improvements occur on TartanDrive (ATE 4.90 → 4.61) and SCAND (RPE 0.67 → 0.51). These results indicate that depth enhances geometric understanding, yielding more physically consistent and reliable trajectories.

\noindent{\textbf{Effect of Trajectory-scale Regularization Loss.}}
The Trajectory-scale Regularization (TSR) loss enforces endpoint alignment with navigation goals, leading to consistent improvements across datasets. As shown in Tab.~\ref{tab5}, TSR alone improves over the baseline, reducing ATE from 2.09 to 2.06 on RECON and from 2.25 to 1.63 on SCAND, with a notable RPE reduction on SCAND (0.67 $\rightarrow$ 0.45). On TartanDrive, TSR substantially decreases ATE from 4.90 to 2.63, highlighting its effectiveness in long-horizon trajectory stabilization. When combined with Depth, performance is further improved, achieving 1.70 $\rightarrow$ 1.01 ATE on RECON and 2.27 $\rightarrow$ 1.12 on SCAND, while also reducing TartanDrive ATE from 4.61 to 1.94, demonstrating consistent complementary gains in both indoor and driving scenarios.

\noindent{\textbf{Effect of Visual-Guided Action Refinement.}}
The Visual-Guided Action Refinement (VGAR) module refines predicted trajectories using visual-spatial constraints from RGBD tokens. With VGAR, both ATE and RPE improve on most datasets; for TartanDrive, ATE reaches 1.55 and RPE 0.68, demonstrating effective trajectory refinement with a lightweight network.

Overall, the ablation studies confirm that each component meaningfully contributes to performance: depth prediction provides spatial grounding, TSR loss ensures goal-aligned trajectories, and VGAR further corrects long-range predictions. Their combination enables robust perception-action coupling, yielding accurate and physically plausible trajectory generation.

\begin{table*}[t]
\caption{Ablation study on RECON, SCAND, and TartanDrive. Lower ATE/RPE and higher PSNR are better. \textbf{Bold} indicates best performance; \underline{underline} indicates second best.}
\label{tab5}
\centering
\resizebox{0.95\textwidth}{!}{
\begin{tabular}{ccc|ccc|ccc|ccc}
\toprule
 \multicolumn{3}{c|}{Module} & \multicolumn{3}{c|}{RECON} & \multicolumn{3}{c|}{SCAND} & \multicolumn{3}{c}{TartanDrive} \\
  Depth & TSR &  VGAR & ATE$\downarrow$ & RPE$\downarrow$ & PSNR$\uparrow$ & ATE$\downarrow$ & RPE$\downarrow$ & PSNR$\uparrow$ & ATE$\downarrow$ & RPE$\downarrow$ &  PSNR$\uparrow$ \\
\midrule
    &Baseline  &  & 2.09 & 0.70 & 16.65 & 2.25 & 0.67 & 15.04 & 4.90 & 1.07 & 17.20 \\
\midrule

 $\checkmark$ & & & 1.70 & 0.47 & \textbf{17.54} & 2.27 & 0.51 & 16.05 & 4.61 & 0.97 & 17.93 \\
 &   $\checkmark$  &    &   2.06 &  0.69 &   16.37 &  1.63 &  0.45   &  15.44 &   2.63&   0.94 &   17.17  \\ 
 $\checkmark$ & $\checkmark$ & & \underline{1.01} & \underline{0.45} & 17.30 

& \textbf{1.12} & \textbf{0.30} & \textbf{16.13} & \underline{1.94} & \textbf{0.53} & \underline{18.12} \\

 $\checkmark$ & $\checkmark$ &  $\checkmark$ & \textbf{0.94} & \textbf{0.43} & \underline{17.31} & \underline{1.15} & \underline{0.34} & \underline{16.11} & \textbf{1.55} & \underline{0.68} & \textbf{18.15} \\
\bottomrule
\end{tabular}}
\end{table*}

\begin{figure}[h]
    \centering
    \includegraphics[width=1.0\textwidth]{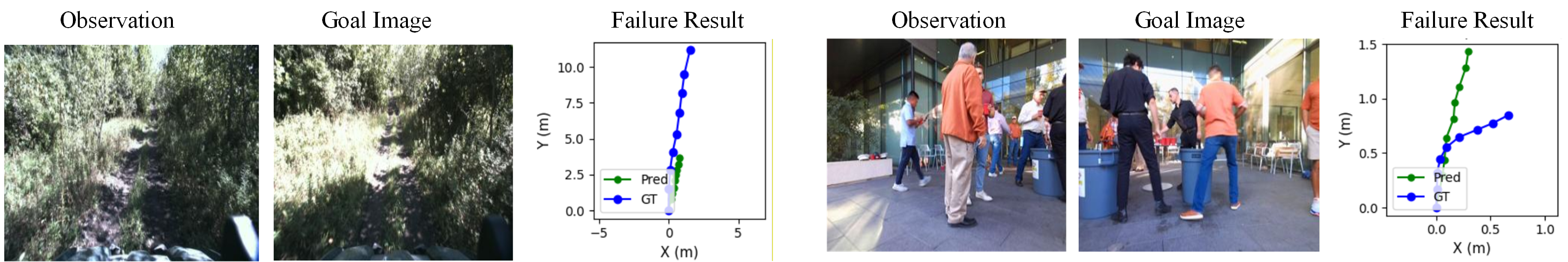}
    \caption{Qualitative results of typical failure scenarios.}
    \label{fig:faliures}
\end{figure}

 \subsection{Failure Cases and Limitation Discussions}

\label{sec:failure}

Despite its robustness, SWAM encounters challenges in scenarios that include ambiguous weeds and sharp turns, as illustrated in Fig.~\ref{fig:faliures}.

The primary failure mode occurs in ambiguous-weeds scenarios, where the model frequently misidentifies traversable weeds as solid obstacles. This perceptual ambiguity, stemming from the lack of semantic-aware traversability reasoning, leads to erroneous avoidance behaviors and accumulated trajectory drift. Another limitation arises in sharp-turn scenarios, where the current planar displacement based action representation fails to capture the rapid heading adjustments required for high-curvature paths. Because the framework does not explicitly encode orientation dynamics, the agent tends to deviate from intended sharp-turn trajectories.

Future work will investigate integrating semantic scene understanding into the model to enable semantic-aware traversability reasoning, as well as developing richer action representations that jointly model position and orientation for more robust long-horizon navigation. Moreover, extending prediction to longer action-video sequences is also a promising direction.




\section{Conclusion}
\label{sec:con}

We presented SWAM, a unified observation–action generation framework for visual navigation that jointly synthesizes intermediate visual observations and corresponding action trajectories. By coupling action and perception in a single-pass inference, SWAM addresses the limitations of candidate-based, verification-centric pipelines, ensuring goal alignment, spatial consistency, and temporal coherence. Our approach leverages spatial priors while requiring only monocular RGB input at inference time. We further propose a visual-guided action refinement module and trajectory-scale regularization loss for improving the action and trajectory prediction performance. Extensive evaluations demonstrate that SWAM surpasses state-of-the-art two-stage baselines in trajectory accuracy, goal success rate, visual fidelity, and inference efficiency, while maintaining stable performance across long-distance and cross-dataset scenarios. These results demonstrate the potential of jointly modeling actions and observations for embodied visual navigation.

\section*{Acknowledgements}
This work was conducted during an internship at Xiaomi EV. We express our sincere gratitude to the entire team at Xiaomi EV for their generous support and collaboration. This work was supported by the National Natural Science Foundation of China under Grant 62371201.

%
%

\bibliographystyle{splncs04}
\bibliography{main}

\title{Pondering the Way: Spatial-perceiving World Action Model for Embodied Navigation 
Supplementary Material
}

\titlerunning{Spatial-Perceiving World Action Model for Navigation}

\author{Hong Chen\inst{1}\orcidlink{0000-0003-1110-713X} \and Daqi Liu\inst{2}\thanks{Project leaders.}\orcidlink{0000-0003-1929-657X} \and Zehan Zhang\inst{2}\protect\footnotemark[1] \and Haiguang Wang\inst{2} \and Tianhao Lu\inst{1} \and Longfei Yan\inst{3} \and Haiyang Sun\inst{2} \and Fangzhen Li\inst{2} \and Hongwei Xie\inst{2} \and Bing Wang\inst{2} \and Guang Chen\inst{2} \and Hangjun Ye\inst{2}\thanks{Corresponding authors.} \and Yihua Tan\inst{1}\protect\footnotemark[2]\orcidlink{0000-0003-0963-5339}   } 



\authorrunning{Chen et al.}

\institute{
Huazhong University of Science and Technology, Wuhan, China \\
\and
Xiaomi EV, Beijing, China \\
\and
Zhejiang University, Hangzhou, China \\
}

\maketitle

\section{Implementation Details}
In the main paper, in order for the model to exhibit navigation capabilities with different planning distances while operating under a fixed sequence length, each training sample consists of a fixed-length segment that contains eight observation–action pairs (9 frames in total).  In addition, we also trained a separate model variant for the long-range visualization analysis in the supplementary material. During training, we randomly sample a segment length $N$ from $\{9, 17, 33, 65\}$, and extract an $N$-frame sub-sequence from the source trajectories. This variable-length training strategy is used only to achieve stable prediction over an extended time range, and it is not used for any of the quantitative comparisons reported in the main paper. To support variable-length sequences, the time-position embeddings use the standard RoPE interpolation rescaling strategy commonly employed in long-context Transformers. This allows the model to maintain consistent time-position encoding across different sequence lengths without modifying the network architecture. Training is conducted using bf16 mixed precision with gradient checkpointing to reduce memory consumption. The model use batch size of 1. We use the Adam optimizer with a learning rate of \(1\times10^{-4}\), a warm-up schedule of 1,000 steps, and gradient clipping with a maximum norm of 1.0. The Adam numerical stability constant is set to \(\epsilon = 1\times10^{-15}\).

\section{More Results}

\begin{table*}[h]
\caption{Depth Estimation Results. Reporting LPIPS.}
\label{depth}
\centering
\begin{tabular}{l|ccc}
\toprule
Datasets  &RECON    & SCAND  &  TartanDrive   \\
\midrule
LPIPS     &  0.224        & 0.243       &0.182 \\
\bottomrule
\end{tabular}
\end{table*}

\begin{figure}[]
    \centering
    \includegraphics[width=1.0\textwidth]{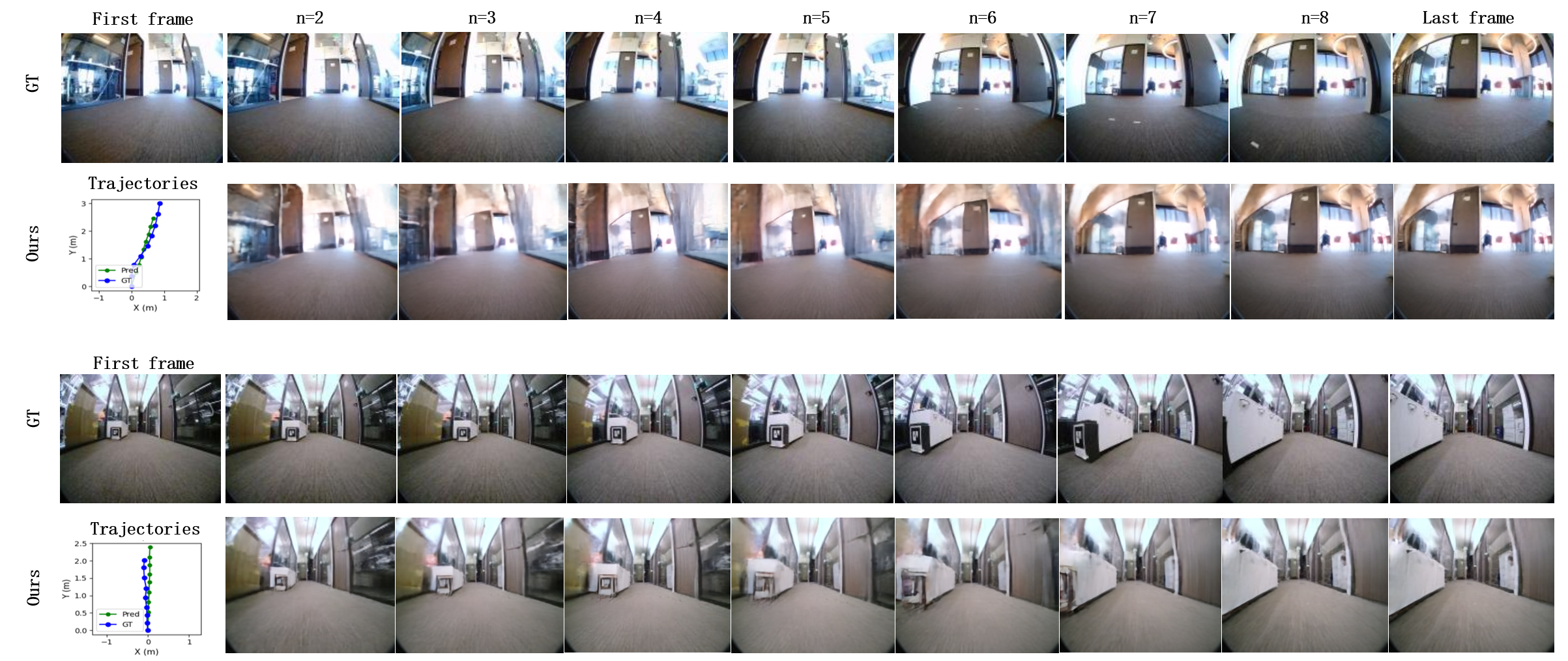}
    \caption{Visualization of zero-shot generalization results on the HuRoN dataset.}
    \label{Exp2_zero}
\end{figure}

\begin{figure}[htbp]
    \centering
    \includegraphics[width=1.0\textwidth]{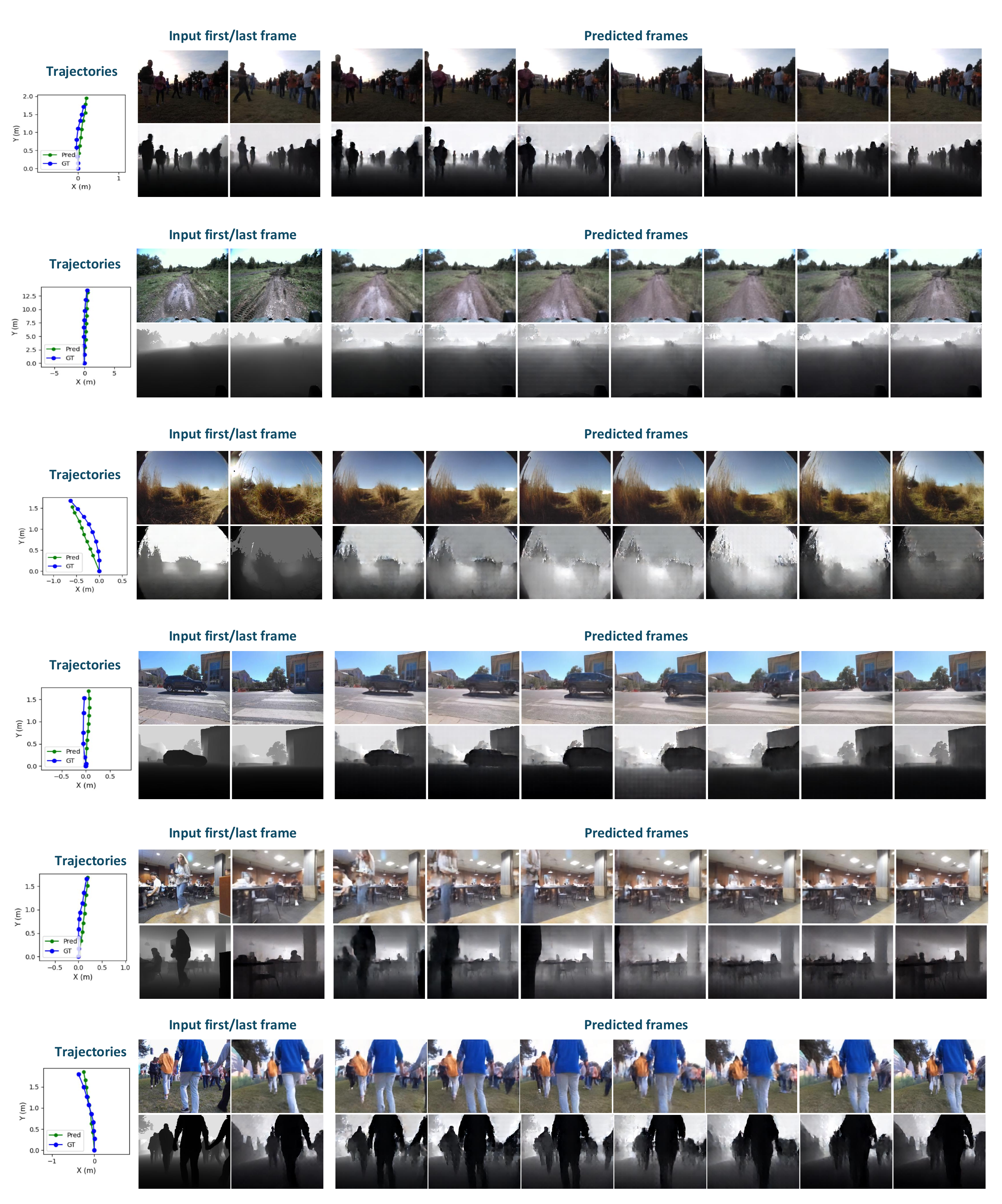}
    \caption{More Visualization results of SWAM.}
    \label{Exp2_all}
\end{figure}

\begin{figure}[htbp]
    \centering
    \includegraphics[width=1.0\textwidth]{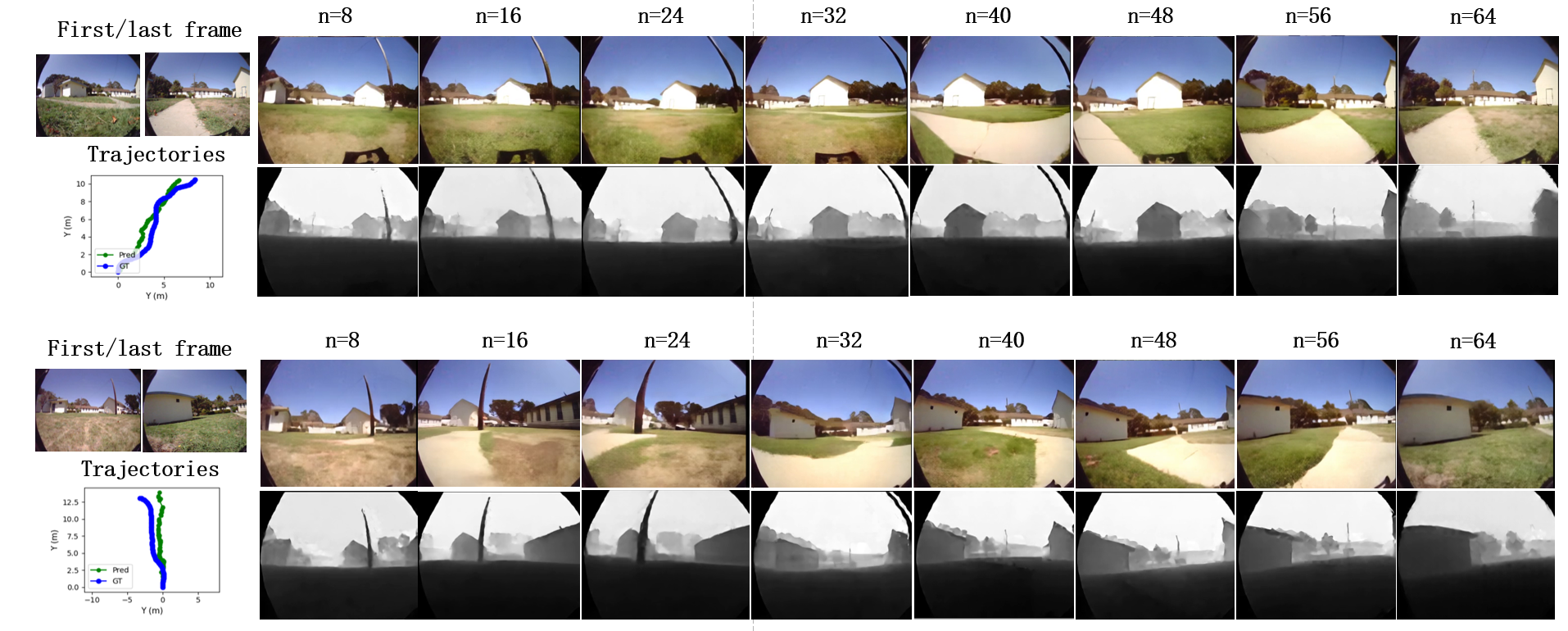}
    \caption{Visualization of long-horizon prediction results of SWAM (64 frames).}
    \label{Exp2_long}
\end{figure}

\subsubsection{Depth Estimation Results.}

We additionally report LPIPS for the generated depth images. 
Since the depth maps used in our experiments are pseudo labels predicted by a pretrained model rather than ground-truth measurements, absolute depth errors may not provide a reliable evaluation. 
Instead, LPIPS is used to measure the perceptual similarity and structural consistency of the generated depth maps. 
As shown in Table~\ref{depth}, SWAM achieves consistently low LPIPS scores across different datasets, indicating that the generated depth maps maintain high perceptual similarity and coherent structural patterns with respect to the reference depth predictions.

\subsubsection{Zero-shot Generalization Visualization Results.} 
We visualize some samples of zero-shot experiments on the public HuRoN dataset, which contains navigation videos captured in previously unseen environments and has a lower spatial resolution compared to the training datasets.

As shown in Fig.~\ref{Exp2_zero}, the predicted videos remain temporally coherent and visually consistent with the observed motion patterns. 
Despite the differences in scene layout, visual appearance, and motion statistics between the training data and the HuRoN dataset, the model is still able to generate stable and realistic future frames. 
These results indicate that SWAM learns transferable spatiotemporal representations and exhibits strong cross-domain generalization capability.

\subsubsection{Longer-sequence Visualization Results.}

We further visualize the model’s ability to perform longer-sequence prediction.  Specifically, the model predicts future frames up to 64 timesteps. As illustrated in Fig.~\ref{Exp2_long}, the generated sequences maintain stable motion dynamics and avoid rapid degradation over time. This indicates that the proposed SWAM enables the model to preserve long-range temporal consistency. Note that a long sequence does not necessarily correspond to a long physical travel distance. This difference is related to factors such as the agent’s movement speed and the data collection efficiency. SWAM, as a high-level planner rather than a local executor, is more concerned with enabling path planning over varying effective distances within a limited computation budget. This capability has been verified in the main paper, and it indicates that the method can adapt robustly to differences across deployed agent embodiments.

\subsubsection{More Visualization Results.}

We provide additional qualitative examples in Fig.~\ref{Exp2_all} to further illustrate the robustness of the proposed method across diverse environments and motion patterns for navigation planning. 
The visualization includes a variety of challenging scenarios, such as low-light environments, open outdoor spaces, regions with dense vegetation or grass, scenes containing dynamic objects, indoor environments, and crowded areas with moving pedestrians. 
Across these different conditions, the generated videos remain temporally coherent while the predicted trajectories follow plausible navigation paths that are consistent with the underlying scene geometry and motion dynamics. 
These results demonstrate that SWAM can effectively capture transferable spatiotemporal representations and maintain stable predictions under significant variations in scene appearance, lighting conditions, and dynamic interactions, highlighting its strong generalization capability across diverse real-world navigation scenarios.

\end{document}